\documentclass[review]{elsarticle}

\usepackage{hyperref}

\usepackage{amsmath,amssymb}
\usepackage[misc,geometry]{ifsym}

\usepackage{booktabs}
\usepackage{multirow}

\usepackage{placeins}

\usepackage{amssymb}
\usepackage{amsmath}
\usepackage{lineno,hyperref}
\modulolinenumbers[5]
\usepackage[table]{xcolor}

\usepackage[]{graphicx}

\journal{Expert Systems with Applications}

\begin{document}

\begin{frontmatter}
	
\title{Leveraging Deep Graph-Based Text Representation for Sentiment Polarity Applications}



\author[mymainaddress]{Kayvan Bijari}
\ead{kayvan.bijari@ut.ac.ir}

\author[mymainaddress]{Hadi Zare\corref{mycorrespondingauthor}}
\cortext[mycorrespondingauthor]{Corresponding author}
\ead{h.zare@ut.ac.ir}

\author[mymainaddress]{Emad Kebriaei}
\ead{emad.kebriaei@ut.ac.ir}

\author[mymainaddress]{Hadi Veisi}
\ead{h.veisi@ut.ac.ir}

\address[mymainaddress]{Faculty of New Sciences and Technologies, University of Tehran, North Kargar Street, Tehran, Iran}

\begin{abstract}
\textcolor{black}{Over the last few years, machine learning over graph structures has manifested a significant enhancement in text mining applications such as event detection, opinion mining, and news recommendation. One of the primary challenges in this regard is structuring a graph that encodes and encompasses the features of textual data for the effective machine learning algorithm. Besides, exploration and exploiting of semantic relations is regarded as a principal step in text mining applications. However, most of the traditional text mining methods perform somewhat poor in terms of employing such relations. In this paper, we propose a sentence-level graph-based text representation which includes stop words to consider semantic and term relations. Then, we employ a representation learning approach on the combined graphs of sentences to extract the latent and continuous features of the documents. Eventually, the learned features of the documents are fed into a deep neural network for the sentiment classification task. The experimental results demonstrate that the proposed method substantially outperforms the related sentiment analysis approaches based on several benchmark datasets. Furthermore, our method can be generalized on different datasets without any dependency on pre-trained word embeddings. }
\end{abstract}
	

\end{frontmatter}



\section{Introduction}
\label{sec:intro}

Text messages are very ubiquitous and they are transferred every day throughout social media, blogs, wikis, news headlines, and other online collaborative media. Accordingly, a prime step in text mining applications is to extract interesting patterns and features from this supply of unstructured data. Feature extraction can be considered as the core of social media mining tasks such as sentiment analysis, event detection, and news recommendation~\citep{aggarwal2018a}.

In the literature, sentiment analysis tends to be used to refer the task of polarity classification for a piece of text at the document, sentence, feature, or aspect level~\citep{liu2012sentiment}. There are various applications on a variety of domains which use sentiment analysis. In this regard, one can mention applying the sentiment analysis for political reviews to estimate the general viewpoint of the parties~\citep{tumasjan2010predicting}, predicting stock market prices based on sentiment analysis by utilizing the different financial news data~\citep{bollen2011twitter}, and making use of the sentiment analysis to recognize the current medical and psychological status for a community~\citep{liu2012sentiment}.

Machine learning algorithms and statistical learning techniques have been rising in a variety of scientific fields~\citep{detmer2018development,eshtay2018improving}. Several machine learning techniques have been proposed to perform sentiment analysis. As one of the powerful sub-domains of machine learning in recent years, deep learning models are emerging as a persuasive computational tool, they have affected many research areas and can be traced in many applications. With respect to the deep learning, textual deep representation models attempt to discover and present intricate syntactic and semantic representations of texts, automatically from data without any handmade feature engineering. Deep learning methods coupled with deep feature representation techniques have improved the state-of-the-art models in various machine learning tasks such as sentiment analysis~\citep{mikolov2013efficient,pennington2014glove} and text summarization~\citep{yousefi2017text}.

Inspired by the recent advances in feature learning and deep learning methods, it is determined that inherent features can be learned from the raw structure of data using learning algorithms. This technique is called representation learning which aids to promote and advance functionality of machine learning methods. To put it differently, representation learning is able to map or convert raw data into a set of features which are considerably more distinctive for machine learning algorithms.

\textcolor{black}{
This research proposes a new approach that takes advantage of graph-based representation of documents integrated with representation learning through the Convolutional Neural Networks (CNN)~\citep{schmidhuber2015deep}. Graph representation of documents reveals intrinsic and deep features compared to the traditional feature representation methods alike bag-of-words (BOW)~\citep{manning1999foundations}. Although words alone play the most important role in determining the sentence's sentiment, their position in a document, as well as their vicinity, can reveal hidden aspects of the sentiment~\citep{Violos2016SentimentAU}. Sometimes the sentiment orientation changes drastically when considering word order. By way of illustration, in the bag-of-words model, the general recommendation is to exclude stop words from the texts. However, stop words can convey meaningful and valuable features for sentiment analysis and their position in the sentence can easily change the polarity of a sentence. With the intention of graph representation, every individual word is depicted as a node in the graph and the interactions between different nodes are modeled through undirected, directed, or weighted edges.\\
In this work, a graph-based representation for text documents is proposed that embodies the textual data at a sentence level.  Afterward, a representation learning on the combined sentence graphs is applied based on a random walker algorithm to fabricate an unsupervised features representation of the documents. 
The well-known deep neural network architecture, CNN, is employed on the learned features of sentiment polarity tasks. \\
While conventional sentiment analysis methods usually ignore stop words, word positions, and orders, our experimental results on benchmark data have justified the significant strength of comprising all of these meaningful elements in a graph-based structure by demonstrating performance gains in sentiment classification.\\
The main contributions of our work are summarized below:
\begin{itemize}
\item We propose an integrated framework for sentiment classification by representing the text document in a graph structure that considers all the informative data.
\item We apply a random walk based approach to learn continuous latent feature representation from the combined graphs of sentences in an unsupervised way.
\item The convolutional neural network is then employed on the vectorized features for sentiment polarity identification without the need for pre-trained word vectors.
\item We demonstrate the usefulness and strength of this integrated graph-based representation learning approach for the sentiment classification tasks based on several benchmark datasets.
\end{itemize} 
}

The overall structure of this paper is as follows. Section \ref{sec:related} begins by reviewing the related works of sentiment analysis and presents the basic idea behind the proposed approach. Section \ref{sec:methodology} discusses the methodology of the proposed method and demonstrates how graph representation and feature learning are used to perform sentiment analysis. A brief introduction of the standard datasets and experimental results of the proposed approach versus some well-known algorithms are given in Section \ref{sec:experimental-results}. Section \ref{sec:conclusion} ends the paper with a conclusion and some insights for future works.

\section{Related Works and Basic Idea}
\label{sec:related}

\subsection{Related Works}

Over the last few years, broad research on sentiment analysis through supervised~\citep{oneto2016statistical}, semi-supervised~\citep{hussain2018semi}, and unsupervised~\citep{garcia2018w2vlda} machine learning techniques have been done. Go et al.~\citep{go2009twitter} were among the firsts who applied distant supervision technique to train a machine learning algorithms based on emoticons for sentiment classification. Researchers in the field of natural language processing carried out a variety of new algorithms to perform sentiment analysis~\citep{taboada2011}. Some distinguished works are further discussed in this section.

As a sub-domain of information retrieval and natural language processing, sentiment analysis or opinion mining can be viewed from different levels of granularity namely, sentence level, document level, and aspect level;  from the point of view of sentence level, Liu's works can be mentioned as one of the pioneers in this field~\citep{hu2004mining}. Works by Pan and Lee can also be considered in which document level of sentiment analysis is examined~\citep{pang2004sentimental}. Lately aspect level of sentiment analysis has attracted more attention, research by Agarwal can be listed in this regard~\citep{agarwal2009contextual}.

Graph-based representation techniques for sentiment analysis have been used in a variety of research works.  \cite{minkov2008learning} considered text corpus as a labeled directed graph in which words represent nodes, and edges denote syntactic relation between the words. They proposed a new path-constrained graph walk method in which the graph walk process is guided by high-level knowledge about essential edge sequences. They showed that the graph walk algorithm results in better performance and is more scalable. In the same way, \cite{violos2016sentiment} suggested the word-graph sentiment analysis approach. In the model, they proposed a well-defined graph structure along with several graph similarity methods, afterward, the model extracts feature vectors to be used in polarity classification. Furthermore, \cite{goldberg2006seeing} proposed a graph-based semi-supervised algorithm to perform sentiment classification through solving an optimization problem, their model suits situations in which data labels are sparse.

Deep learning methods are operating properly on the field of sentiment analysis.  A semi-supervised approach was introduced in \cite{socher2011semi} based on recursive autoencoders to foresee sentiment of a given sentence. The system learns vector representation for phrases and exploits the recursive nature of sentences. They have also proposed a matrix-vector recursive neural network model for semantic compositionality. It can learn compositional vector representations for expressions and sentences with discretionary length ~\citep{socher2012semantic}. To clarify, the vector model catches the intrinsic significance of the component parts of sentences, while the matrix takes the importance of neighboring words and expressions into account.  Recursive neural tensor network (RNTN) was proposed to represent a phrase through word vectors and a parse tree ~\citep{socher2013recursive}. Their model computes nodes vectors in a tree-based composition function.

Other deep architectures have been applied for natural language processing tasks~\citep{chen2017improving}. The semantic role labeling task is investigated by employing convolutional neural networks \citep{collobert2011natural}.  In another attempt,~\cite{collobert2011deep} exploited a convolutional network with similar architecture that serves syntactic parsing. In addition,~\cite{poria2016convolutional} applied a convolutional neural network to extract document features and then employed multiple-kernel learning (MKL) for sentiment analysis. In another work, \cite{poria2017context} a long short-term memory network was used to extract contextual information from the surrounding sentences.

Unlike deep learning methods, which use neural networks to transform feature space into high dimensional vectors, general practices for sentiment analysis take advantage of basic machine learning methods. Indeed,~\cite{tripathy2016classification} ensembles a collection of machine learning techniques along with n-grams to predict sentiment of a document. Additionally, evolutionary algorithms have been utilized for several optimization problems~\citep{bijari2018memory}, ALGA~\citep{keshavarz2017alga} makes use of evolutionary computation to determine optimal sentiment lexicons which leads to a better performance.

\subsection{Motivation}

In natural language processing, bag-of-word representation is one of the most common means to represent the features of a document. However, it is insufficient to describe the features of a given document due to several limitations such as lacking word relations, scalability issues, and neglecting semantics~\citep{gharavi2016deep}. To mitigate these shortcomings, some other representation techniques are proposed to model textual documents~\citep{tsivtsivadze2006locality}. These methods can take into account a variety of linguistic, semantic, and grammatical features of a document.

The decency of solutions that a machine learning algorithms provide for a task such as classification, heavily depends on the way of features representation in the solution area. Different feature representations techniques can entangle (or neglect) some unique features behind the data. This is where feature selection and feature engineering methods come into play and seek to promote and augment the functionality of machine learning algorithms~\citep{zare2016relevant}.

Feature engineering methods accompanying domain-specific expertise can be used to modify basic representations and extract explanatory features for the machine learning algorithms. On the other hand, new challenges in data presentation, advancements in artificial intelligence, and probabilistic models drive the need for representation learning techniques and feature learning methods. Feature learning can be defined as a transformation of raw data input to a new representation that can be adequately exploited in different learning algorithms~\citep{bengio2013representation}.

As indicated previously in the introduction, the main idea of the proposed method is to shape sentences in a document as a graph, afterward analyze the graphs utilizing network representation learning approaches. accordingly, the proposed method entails three main phases namely, graph representation, feature learning, and classification. A detailed description of the components of the proposed method is specified in Section~\ref{sec:methodology}.

\FloatBarrier
\section{Elements of the proposed method}
\label{sec:methodology}

\begin{figure}[!ht]
	\centering
	\includegraphics[width=0.85\textwidth]{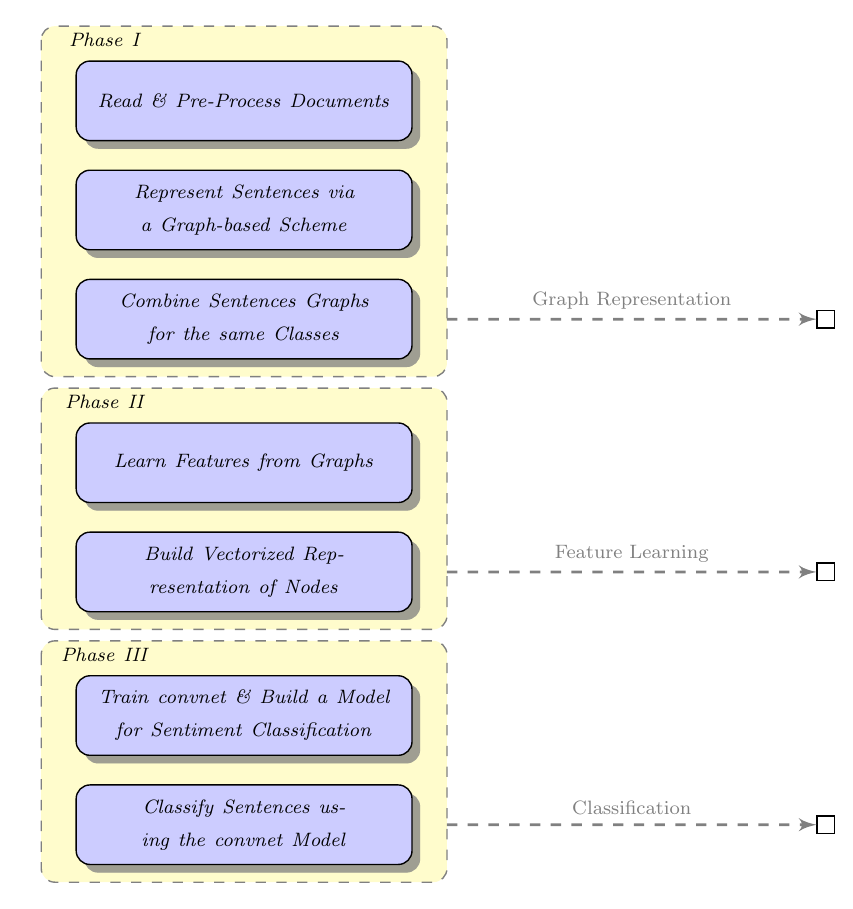}
	\caption{work-flow of the proposed sentiment classification approach}
	\label{fig:work-flow}
\end{figure}

The proposed method is comprised of three primary building blocks which will be later explained. Initially, textual documents are pre-processed and then transformed into word-graphs. Afterward, through a feature learning technique (representation learning phase), inherent and intrinsic features of the word-graphs are determined. In the end, a convolutional neural network is trained based on the extracted features and employed to perform the sentiment classification task. Figure \ref{fig:work-flow} illustrates the work-flow of the proposed method.

\subsection{Graph Representation}
\label{subsec:graph-representation}

In the era of big data, text is one of the most ubiquitous forms of storing data and metadata. Data representation is the vital step for the feature extraction phase in data mining. Hence, a proper text representation model which can considerably picture inherent characteristics of textual data, is still an ongoing challenge. Because of simplicity and shortcomings of traditional models such as the vector space model, offering new models is highly valued. Some disadvantageous of classical models such as bag-of-words model can be listed as follows~\citep{gharavi2016deep}: 

\begin{itemize}
\item Meaning of words in the text and textual structure cannot be accurately represented.
\item Words in the text are considered independent from each other.
\item Word's sequences, co-occurring, and other relations in a corpus is neglected.
\end{itemize}

Broadly speaking, words are organized into clauses, sentences, and paragraphs in order to describe the meaning of a document. Moreover, their occurring, ordering, and positioning, as well as the relationship between different components of the document are necessary and valuable to understand the document in detail.

Graph-based text representation can be acknowledged as one of the genuine solutions for the aforementioned deficiencies. A text document can be represented as a graph in many ways. In a graph, nodes denote features and edges outline the relationship among different nodes. Although there exist various graph-based document representation models~\citep{violos2016sentiment}, the co-occurrence graph of words is an effective way to represent the relationship of one term over the other in the social media contents such as Twitter or short text messages. The co-occurrence graph is called word-graph in the rest of the paper. 

Word-graph is defined as follows: given a sentence $ S $, let $ W $ be the set of all words in the sentence $ S $. A Graph $ G(E,W) $ is constructed such that any $ w_i, w_j \in W $ are connected by $ e_k \in E, \quad if \quad \exists R_l \quad s.t. \quad R_l(w_i, w_j) \in R $. 

\begin{figure}[!ht]
	\centering
	\includegraphics[width=0.85\textwidth]{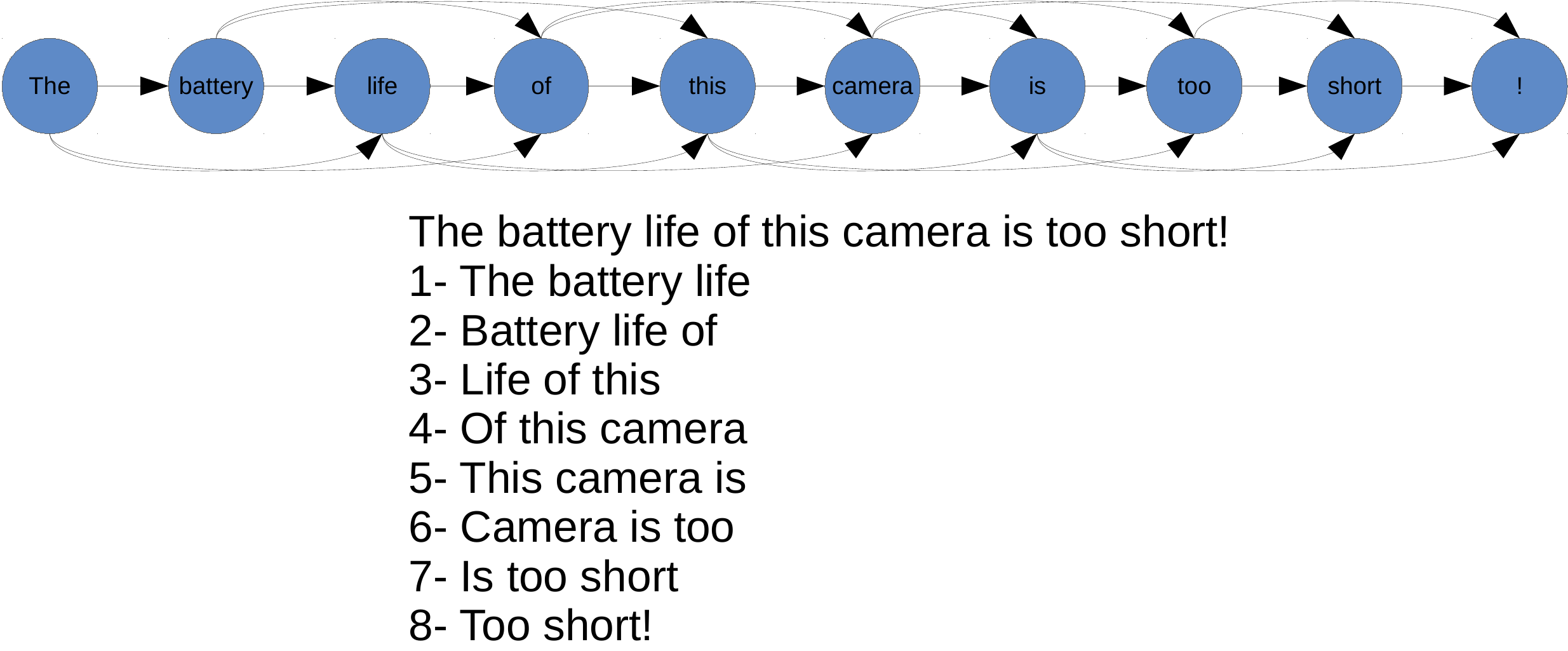}
	\caption{A sample sentence graph with word-window 3, and sub-sentences which each window give importance to.}
	\label{fig:sent-graph}
\end{figure}

In other words, in the graph $ G $ any word in the sentence is considered as a single vertex. Two vertices are connected by the edge $ e_k $, if there exists a connection between them governed by the relation $ R $.
The relation $ R $ is satisfied if, for instance, its corresponding lexical units co-occur within a window of maximum $ N $ words, where $ N $ can be set to any integer (typically, 2 to 10 seems to be fine based on different trials). The proposed method uses word graphs with window of size 3. Figure \ref{fig:sent-graph} presents a graph of a sample sentence with word-window of size 3. Relation $ R $ in this graph is satisfied when two nodes are within a window with a maximum length of 3.

\subsection{Feature Learning}
\label{subsec:feature-learning}

To perform well on a given learning task, any (un)supervised machine learning algorithm requires a set of informative, distinguishing, and independent features. One typical solution is to feed the algorithms with hand-engineered domain-specific features based on human ingenuity and expert knowledge. However, feature engineering designates algorithm's lack of efficiency to entangle and organize the discriminative features from the data. Moreover, feature engineering not only requires tedious efforts and labor, but it is also designed for specific tasks and can not be efficiently generalized across other tasks~\citep{grover2016node2vec}. Because of the broad scope and applicability of machine learning algorithms for different jobs, it would be much beneficial to make machine learning algorithms less dependent on feature engineering techniques.

An alternative to feature engineering is to enable algorithms to learn features of their given task based on learning techniques. As one of the new tools in machine learning, representation learning and feature learning enables machines and algorithms to learn features on their own directly from data. In this regard, features are extracted by exploiting learning techniques and making transformation on raw data for the task. Feature learning allows a machine to learn specific tasks as well as it features and obviates the use of feature engineering~\citep{bengio2013representation}. 

Node embedding is a vectorized representation of nodes for each graph. It is trained via feature learning algorithms so that to pay more attention to the important nodes and relations while ignoring the unimportant ones. More specifically, in the proposed method as a novel feature learning algorithm, node2vec, is used to reveal the innate and essential information of a text graph ~\citep{grover2016node2vec}. Afterward, a convolutional neural network is used to learn and classify text graphs.

Node2vec~\citep{grover2016node2vec} together with Deepwalk~\citep{perozzi2014deepwalk} and LINE~\cite{tang2015line} are well-known algorithms for representation learning on the graph structure. The main goal of such algorithms is to pay more attention to the important nodes and relations while paying less to the unimportant ones. In other words, a feature learning algorithm is used to reveal the innate and essential information of a graph.

Node2vec is a semi-supervised algorithm which is intended for scalable feature learning in graph networks. The purpose of this algorithm is to optimize a graph-based objective function through stochastic gradient descent. By making use of a random walker to find a flexible notation of neighborhoods, node2vec returns feature representations (embeddings) that maximize the likelihood of maintaining the graph's structure (neighborhoods)~\citep{grover2016node2vec}. In the proposed method representation learning is done based on the node2vec framework by virtue of its scalability and effectiveness in exploring graph networks as compared to other algorithms. The work-flow of the feature learning in the proposed algorithm is further discussed in the following paragraphs.

Feature learning in networks is formulated as a maximum likelihood optimization problem. Let $G = (V, E)$ be a given (un)directed word-graph. Let $f: v \rightarrow R^d$ be the mapping function from nodes to feature representation which is to be learned for a distinguished task. $d$ is a parameter which designates the number of dimensions of the feature to be represented. Equivalently, $f$ is a matrix of size $|v| \times d$ parameters. For every node in the graph $ u \in V$, a neighborhood $ N_{S(u)} \subset V $ is defined. 

The following optimization function which attempts to maximize the log-probability of observing neighborhood $N_S(u)$ for node $u$, is defined as equation~\eqref{eq:objective}. 

\begin{equation}
\label{eq:objective}
\max_{f} \sum_{u\in V} log(P(N_S(u)|f(u)))
\end{equation}

To make sure that the equation~\eqref{eq:objective} is tractable, two standard assumptions need to be made.
\begin{itemize}
\item Conditional independence. Likelihood is factorized in such a way that the likelihood of observing a neighborhood node is independent of observing any other neighborhood. According to this assumption, $P(N_S(u)|f(u))$ can be rewritten as equation~\eqref{eq:cond-ind}. 
\begin{equation}
\label{eq:cond-ind}
P(N_S(u)|f(u)) = \prod_{n_i \in N_S(u)} P(n_i|f(u))
\end{equation}

\item Symmetry in feature space. A source node and neighborhood node have a symmetric impact on each other. Based upon this assumption $P(n_i|f(u))$ is calculated using equation~\eqref{eq:symmetry} in which conditional likelihood of every source-neighborhood node pair can be modeled as a softmax unit parametrized by the dot product of their features.

\begin{equation}
\label{eq:symmetry}
P(n_i|f(u)) = \frac{exp(f(n_i) \cdot f(u))}{\sum_{v\in V}exp(f(v) \cdot f(u)}
\end{equation}
\end{itemize}

Based on these two assumptions, the objective function in equation~\eqref{eq:objective} can be simplified, 
\begin{equation}
\label{eq:objective-simplified}
\max_{f} \sum_{u\in V} \Bigl[ -log(Z_u) \quad + \sum_{n_i \in N_S(U)}f(n_i)\cdot f(u) \Bigl]
\end{equation}
where, per-node partition function, $ Z_u = \sum_{u\in V}exp(f(u)\cdot f(v) $. Equation~\eqref{eq:objective-simplified} is then optimized using stochastic gradient ascent over model parameters defining the features $f$.

The neighborhoods $N_S(u)$ are not restricted to direct neighbors, but it is generated using sampling strategy $S$. There are many search strategies to generate neighborhood $N_S(u)$ for a given node $u$, simple strategies include breadth-first sampling which samples immediate neighbors, and depth-first sampling which seeks to sample neighbors with the most distant from the source. For a better exploration of the graph structure, a random walk manner is used as a sampling strategy which smoothly interpolates between BFS (Breadth First Search) and DFS (Depth First Search) strategies~\cite{manber1989introduction}. In this regard, given a source node $u$, a random walk of length $l$ is simulated. Let $c_i$ be the $i$-th node in the walk, starting with $c_0=u$. Other nodes in the walk, are generated using the following equation~\eqref{eq:walk}.

\begin{equation}
\label{eq:walk}
P(c_i = x | c_{i-1} = v) = 
\begin{cases}
\frac{\pi_{vx}}{Z} & \text{if $(v, x) \ in $  $ E $,} \\
0 & \text{otherwise.}
\end{cases}
\end{equation}
where $\pi_{vx}$ is the transition probability between given nodes $v$ and $x$, and $Z$ is the normalizing constant.

Given $W$ is (un)weighted adjacency matrix of the graph, $v$ is the node that random walk is resides at, and $t$ to be the traversed edge. The transition probability matrix $\pi$ is defined as $\pi_{vx} = \alpha_{pq}(t,x) \times W_{vx}$. $\alpha_{pq}$ is calculated using the following equation~\eqref{eq:trans}.

\begin{equation}
\label{eq:trans}
\alpha_{pq}(t,x) = 
\begin{cases}
\frac{1}{p} & \text{if $ d_{tx} = 0 $} \\
1 & \text{if $ d_{tx} = 1 $} \\
\frac{1}{q} & \text{if $ d_{tx} = 2 $}
\end{cases}
\end{equation}

where $d_{tx}$ is the shortest path distance between nodes $t$ and $x$ and its value should be one of $\{0, 1, 2\}$. Values $p$ and $q$ are control parameters that determine how fast or slow the walk explores the neighborhood of the starting node $u$. They allow the search procedure to interpolate between BFS and DFS to investigate different notions of neighborhoods for a given node.

\subsection{ConvNet Sentiment Classification}
\label{subsec:convnet-based-classification}

Adopted from neurons of the animal's visual cortex, ConvNets or convolutional neural networks is a biologically inspired variant of a feed-forward neural network~\citep{schmidhuber2015deep}. ConvNets have shown to be highly effective in many research areas such as image classification and pattern recognition tasks~\citep{sharif2014cnn}. They have also been successful in other fields of research such as neuroscience~\citep{gucclu2015deep} and bioinformatics~\citep{ji20133d}.

Similar to the general architecture of neural networks, ConvNets are comprised of neurons, learning weights, and biases. Each neuron receives several inputs, takes a weighted sum over them, passes it through an activation function at its next layer and responds with an output. The whole network contains a loss function to direct the network through its optimal goal, All settings that will apply on the basic neural network~\citep{goodfellow2016deep}, is likewise applicable to ConvNets.

Apart from computer vision or image classification, ConvNets are applicable for sentiment and document classification. Inputs for the deep algorithms are sentences or documents which are represented in the form of a matrix such that each row of the matrix corresponds to one token or a word. Besides, each row is a vectorized representation of the word and the whole matrix will represent a sentence or a document. In deep learning based approaches, these vectors are low-dimensional word embedding resulted from algorithms and techniques such as word2vec~\citep{mikolov2013efficient}, GloVe~\citep{pennington2014glove}, or FastText~\cite{joulin2016fasttext}

In the proposed method, a slight variant of ConvNet architecture of Kim~\citep{kim2014convolutional} and Collobert~\citep{collobert2011natural} is used for sentiment classification of sentences. Let $n_i \in \mathbb{R}^d $ the d-dimensional node embedding corresponding to $i$-th node in a word-graph of a given document. It should be noted that sentences are padded beforehand to make sure that all documents have the same length. 

For convolution operation, a \textit{filter} $ w \in \mathbb{R}^{xd}$ is applied on nodes to produce a new feature, $c_i$ in equation~\eqref{eq:filter}, form a set of $x$ nodes.

\begin{equation}
\label{eq:filter}
c_i = f(w \cdot n_{i:i+x-1} + b)
\end{equation}

Where $b$ is bias term and $f$ is a non-linear function such as hyperbolic tangent. This filter is applied to any possible nodes in the graph $\{g_1, g_2, g_3, \cdots , g_x \}$ to create the \textit{feature map} $c$ in equation~\eqref{eq:feature-map}. 

\begin{equation}
\label{eq:feature-map}
c = [c_1, c_2, c_3, \cdots, c_{x}]
\end{equation}

Afterwards, a max-over-time pooling operation~\citep{collobert2011natural} is performed over the feature map and takes maximum value, $ \hat{c} = \max\{c\} $, as a feature corresponding to this particular filter. This idea is to capture and keep the most important features for each estimated map. Furthermore, this max-pooling deals with the documents with an uncertain length of sentences which were padded previously.

\begin{figure}[!t]
	\centering
	\includegraphics[width=.85\textwidth]{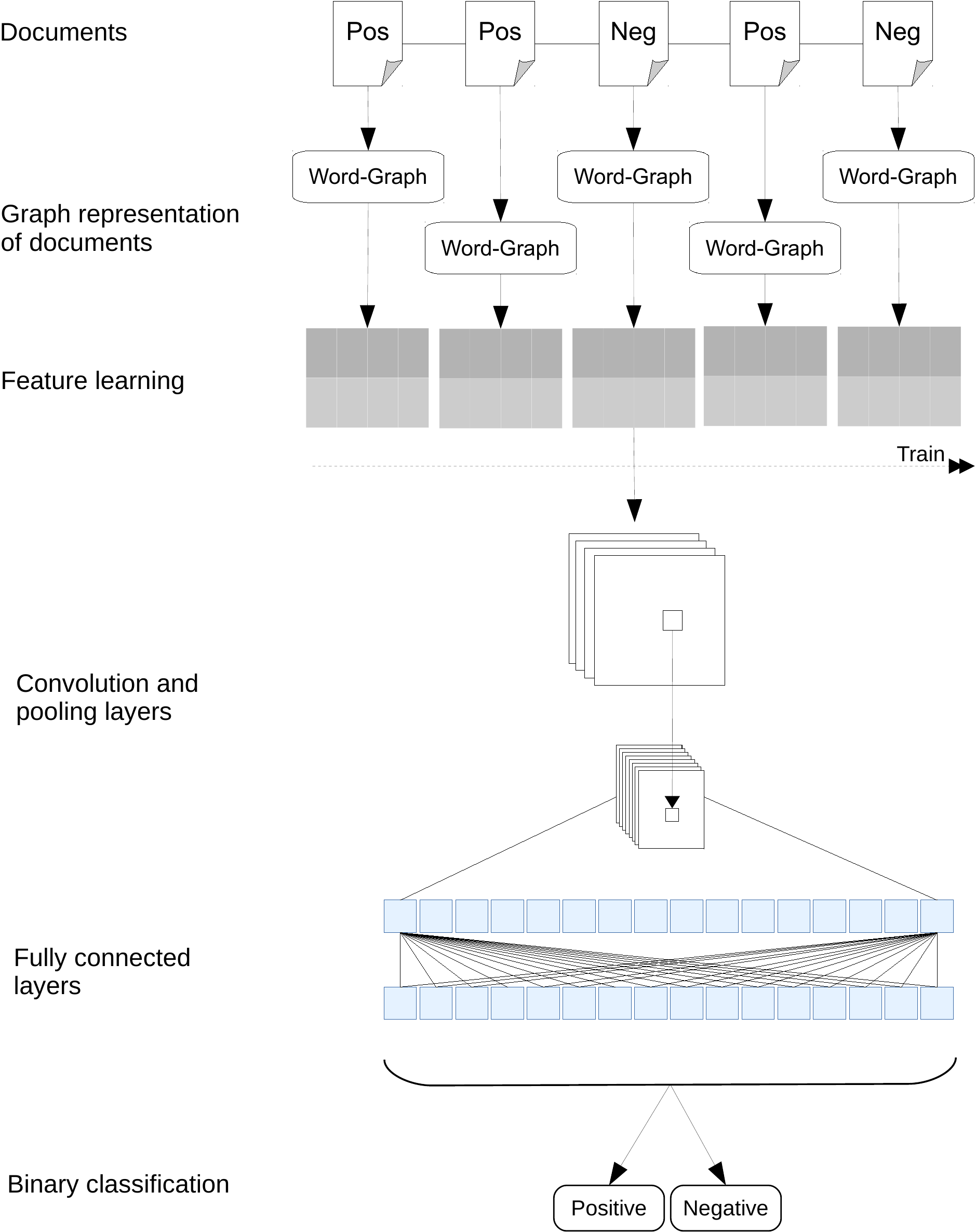}
	\caption{The model architecture of a multi-channel CNN network for sample documents. First, documents are converted into word-graphs. Then, using a feature learning algorithm, node2vec, structure of the graph is transformed into a set of meaningful features. Afterward, via convolution and max-pooling layers, CNN learns the distinguishing features of each document, and a fully connected softmax layer performs the sentiment classification.}
	\label{fig:cnn-model}
\end{figure}

The above description was a procedure in which a feature is extracted from a single filter. The ConvNet model utilizes multiple features each with varying window-sizes to extract diverse features. Eventually, these features fabricate next to the last layer and are passed into a fully connected softmax layer which yields the likelihood probability over the sentiment labels. Figure~\ref{fig:cnn-model} reveals the architecture of the proposed method accompanying its different parts.

\FloatBarrier
\section{Experimental Results}

\label{sec:experimental-results}
This section is devoted to the experimental results of the proposed method on a set of public benchmark datasets for sentiment classification. First, an introduction to the benchmark datasets and some statistics is given. Then, the performance of the proposed method would be evaluated compared to some well-known machine learning techniques. 

\subsection{Datasets}
\label{sec:datasets}
An essential part of examining a sentiment analysis algorithm is to have a comprehensive dataset or corpus to learn from, and a test dataset to make sure that the accuracy of your algorithm meets the expected standards. The proposed method was investigated on different datasets which are taken from Twitter and other well-known social networking sites. These datasets are ``HCR'',`` Stanford'', ``Michigan'', ``SemEval'', and ``IMDB''. These datasets are briefly introduced in the following. 

\subsubsection{Health-care reform (HCR)}
The tweets of this dataset are collected using the hash-tag ``\#hcr'' in March 2010~\citep{speriosu2011twitter}. In this corpus, only the tweets labeled as negative or positive are considered. This dataset consists of 1286 tweets, from which 369 are positive and 917 are negative.

\subsubsection{Stanford}
The Stanford Twitter dataset was originally collected by Go et al.~\citep{go2009twitter} this test dataset contains 177 negative and 182 positive tweets.

\subsubsection{Michigan}
This data set was collected for a contest in university of Michigan. In this corpus each document is a sentence extracted from social media or blogs, sentences are labeled as positive or negative. The Michigan sentiment analysis corpus contains totally 7086 sentences which  3091 samples are negative and 3995 positive samples. 

\subsubsection{SemEval}
The SemEval-2016 corpus~\citep{nakov2016semeval} was built for Twitter sentiment analysis task in the Semantic Evaluation of Systems challenge (SemEval-2016). 14247 tweets were retrieved for this dataset, of which 4094 tweets are negative and the rest 10153 tweets categorized as positive.

\subsubsection{IMDB}
10,000 positive, 10,000 negative full text movie reviews. Sampled from original Internet movie review database of movies reviews. Table \ref{tbl:dataset} briefly summarizes the datasets which are being used for evaluation of the proposed method.

\begin{table}[ht]
	\centering
	\caption{Distribution of negative, positive samples in the given datasets, which will be used for evaluation.}
	\label{tbl:dataset}
	\begin{tabular}{lccccc}
		\toprule
		Dataset & HCR & Stanford & Michigan & SemEval & IMDB \\
		\midrule
		
		Positive & 369  & 182  & 3995 & 10153 & 10,000 \\
		Negative & 917  & 177  & 3091 & 4094  & 10,000 \\
		Total    & 1286 & 359  & 7086 & 14247 & 20,000 \\
		
		\bottomrule
	\end{tabular}
\end{table}

\subsection{Evaluation Metrics}

Sentiment analysis can be viewed as a classification problem. Therefore, the well-known information retrieval (IR) metrics can be used to evaluate the results of the sentiment analysis algorithms. Most of the evaluation metrics are based on the calculation of the values such as TP, TN, FP, and FN which can be used to form the confusion matrix to describe the performance of a classification model, see \cite{manning2008} for further details. 

Table~\ref{tbl:eval} describes accuracy, precision, recall, and F1 score which are applied to assess and evaluate classification algorithms. 
\begin{table}[ht]
	\centering
	\caption{Evaluation metrics for classification algorithms}
	\label{tbl:eval}
	\begin{tabular}{lc}
		\toprule
				
		Evaluation metric & Mathematical definition \\ 
		
		\midrule
		Accuracy & $\frac{TP + TN}{TP+TN+FP+FN}$\\
		Precision & $\frac{TP}{TP+FP}$\\
		Recall & $\frac{TP}{TP+FN}$\\
		F1 & $\frac{2 \times precision \times recall}{precision + recall}$ \\
		
		\bottomrule
	\end{tabular}
\end{table}

\subsection{Compared Algorithms}

The proposed method is challenged against well-known classification algorithms that is discussed in the following.

Support vector machine (SVM) is a supervised machine learning algorithm which performs a non-linear classification using kernel idea to implicitly transform the data into a higher dimension. Data is then inspected for the optimal separation boundaries, between classes. In SVMs, boundaries are referred to as hyperplanes, which are identified by locating support vectors or the instances that define the classes. Margins, lines parallel to the hyperplane, are defined by the shortest distance between a hyperplane and its support vectors. Thereupon, SVMs can classify both linear and nonlinear data. In general terms, SVMs are very beneficial when there is a huge number of features in cases such as text classification or image processing. The grand idea with SVMs is, with enough number of dimensions, a hyperplane separating a particular class from all others can always be found. Essentially, SVMs looks not just for any separating hyperplane but the maximum-margin hyperplane which remains at the equal distance from respective class support vectors~\cite{michalski2013machine}. Due to high dimensionality, sparsity, and linearly separability in the feature space of textual documents linear kernel is a decent choice for text classification with SVMs~\cite{leopold2002text}. Besides, it is shown that the choice of kernel function does not affect text classification performance much~\cite{joachims1998text}. In this paper, we used SVM with linear (C=1.00) and RBF (C=100, $\gamma$=0.1) kernels, the implementations for this purpose are taken from sklearn package written in python 2.7~\cite{pedregosa2011scikit}.

Naive Bayes classifiers are probabilistic classifiers that are known to be simple and yet highly efficient. The probabilistic model of naive Bayes classifiers is based on Bayes’ theorem, and the adjective naive alludes the assumption that features in a dataset are mutually independent. in practical terms, the independence assumption is often violated, however, naive Bayes classifiers still perform adequately and can outperform the other compelling alternatives. Here we use term frequency-inverse document frequency (Tf-idf) with Naive Bayes for sentiment classification. The Tf-idf is a technique for characterizing text documents. It can be interpreted as a weighted term frequency, it assumes that the importance of a word is inversely proportional to how often it occurs across all documents. Although Tf-idf is most commonly employed to rank documents by relevance in different text mining tasks such as page ranking, it can also be utilized to text classification through naive Bayes~\cite{liu2017research}.

 Convolutional neural network is employed on the experiments to compare with the proposed approach. This network typically includes two operations, which can be considered of as feature extractors, convolution and pooling. CNN performs a sequence of operations on the data in its training phase and the output of this sequence is then typically connected to a fully connected layer which is in principle the same as the traditional multi-layer perceptron neural network. More detail about this type of network is given in section~\ref{subsec:convnet-based-classification}. Other hyperparameters for the CNN model as well as the one which is used in the proposed method are shown in the~\ref{tbl:hyper}.

\begin{table}[ht]
	\centering
	\caption{Hyperparameters of the CNN algorithms}
	\label{tbl:hyper}
	\begin{tabular}{lc}
		\toprule
		
		Parameter & Value\\ 
		
		\midrule
		Sequence length & 2633\\
		Embedding dimensions & 20\\
		Filter size & (3, 4)\\
		Number of filters & 150 \\
		Dropout probability & 0.25\\
		Hidden dimensions & 150 \\

		\bottomrule
	\end{tabular}
\end{table}

Recursive neural tensor networks (RNTNs) are neural networks useful for natural language processing tasks, they have a tree structure with a neural network at each node. RNTNs can be used for boundary segmentation to determine which word groups are positive and which are not, this can be leveraged to sentences as a whole to identify its polarity. RNTNs need external components like Word2vec, vectors are used as features and serve as the basis of sequential classification. They are then grouped into sub-phrases, and the sub-phrases are combined into a sentence that can be classified by sentiment~\cite{socher2013recursive}. 

\subsection{Results}

We compare performance of the proposed method to support vector machine and convolutional neural network for short sentences by using pre-trained Google word embeddings~\citep{kim2014convolutional}. Table \ref{tbl:imdb-twitter-resutls} presents the results of the different methods and indicates the superiority of the proposed method over its counterparts in most of the cases. It is important to note how well an algorithm is performing on different classes in a dataset, for example, SVM is not showing good performance on positive samples of Stanford dataset which is probably due to the sample size and therefore the model is biased toward the negative class. On the other hand, F1 scores of the proposed method for both positive and negative classes show how efficiently the algorithm can extract features from different classes and do not get biased toward one of them.

\begin{table}[ht]
	\tiny
	\caption{Experimental results on given datasets}
	\label{tbl:imdb-twitter-resutls}
	\resizebox{1\textwidth}{!}{
	\begin{tabular}{lcccccccccc}
		\toprule
		
		\multirow{2}{*}{Method} &
		\multicolumn{3}{l}{Negative class (\%)}  &&
		\multicolumn{3}{l}{Positive class (\%)}  && 
		\multicolumn{2}{l}{Overall (\%)} \\ \cmidrule{2-4} \cmidrule{6-8} \cmidrule{10-11}
		
		& {precision} & {recall} & {F1} &&
		{precision} & {recall} & {F1}  &&
		{accuracy} & {F1} \\ 
		
		\midrule
		\textbf{HCR} &&&&&& \\
		\rowcolor{lightgray} Proposed method(CNN$+$Graph) & 89.11 & 88.60 & 81.31 && 85.17 & 84.32 & 84.20 && 85.71 & 82.12  \\
		SVM(linear) & 80.21 & 91.40 & 85.01 && 67.12 & 45.23 & 54.24 && 76.01 & 76.74  \\
		SVM(RBF) & 77.87 & 99.46 & 87.35 && 95.65 & 29.73 & 45.36 && 79.45 & 45.36  \\
		NB(tf-idf) & 74.04  & 88.00  & 80.42  && 58.00  & 34.94  & 43.61  && 70.93  & 43.60   \\
		Kim(CNN$+$w2v) & 75.39 & 78.69 & 77.71 && 40.91 & 36.49 & 38.52 && 66.53 & 65.94 \\ 
		RNTN(\cite{socher2013recursive}) & 88.64 & 85.71 & 87.15 && 68.29 & 73.68 & 70.89 && 82.17 & 70.88 \\ \hline\hline
		
		\rule{0pt}{3ex}\textbf{Stanford} &&&&&& \\
		\rowcolor{lightgray} Proposed method(CNN$+$Graph) & 86.38 & 90.37 & 91.29 && 77.46 & 56.45 & 65.52 && 83.71 & 78.72  \\
		SVM(linear) & 79.21 & 100.0 & 88.40 && 00.00 & 00.00 & 00.00 && 79.20 & 70.04  \\
		SVM(RBF) & 63.64 & 85.37 & 72.92 && 64.71 & 35.48 & 45.83 && 63.88 & 45.83  \\
		NB(tf-idf) & 61.29  & 54.29  & 57.58  &&  60.98 & 67.57  &  64.10 && 61.11  &  64.10  \\
		Kim(CNN$+$w2v) & 79.96 & 99.59 & 88.70   && 22.22 & 0.56 & 0.95 && 79.72 & 71.10 \\ 
		RNTN(\cite{socher2013recursive}) & 64.29 & 61.36 & 62.79   && 71.33 & 73.82 & 72.55 && 68.04 & 72.54 \\ \hline\hline
		
		\rule{0pt}{3ex}\textbf{Michigan} &&&&&& \\
		\rowcolor{lightgray} Proposed method(CNN$+$Graph) & 98.89 & 98.75 & 98.41 && 98.82 & 98.14 & 98.26 && 98.41 & 98.73   \\
		SVM(linear) & 99.51 & 91.51 & 97.50 && 98.56 & 98.14 & 99.62 && 98.73 & 98.72   \\
		SVM(RBF) & 76.02 & 73.67 & 74.83 && 66.40 & 69.13 & 67.74 && 71.72 & 67.73   \\
		NB(tf-idf) & 76.92  & 74.07  &  75.47 &&  84.78 &  86.67 & 85.71  && 81.94  & 85.71   \\
		Kim(CNN$+$w2v) & 95.64 & 93.43 & 94.58 && 95.12 & 96.73 & 95.46 && 95.31 & 95.34   \\
		RNTN(\cite{socher2013recursive}) & 93.19 & 95.61 & 94.38 && 96.57 & 94.65 & 95.60 && 95.06 & 95.59   \\ \hline\hline
		
		\rule{0pt}{3ex}\textbf{SemEval} &&&&&& \\
		\rowcolor{lightgray} Proposed method(CNN$+$Graph) & 90.80 & 80.35 & 84.81 && 87.32 & 92.24 & 90.76 && 87.69 & 87.78   \\
		SVM(linear) & 77.91 & 61.97 & 69.06 && 85.74 & 92.89 & 89.17 && 83.95 & 83.36   \\
		SVM(RBF) & 24.21 & 30.67 & 27.06 && 72.63 & 65.71 & 69.00 && 56.49 & 69.00   \\
		NB(tf-idf) & 28.57  & 23.53  & 25.81  && 77.59  & 81.82  & 79.65  && 68.05  & 79.64   \\
		Kim(CNN$+$w2v) & 57.87 & 42.26 & 46.97 && 78.85 & 85.13 & 81.87 && 72.50 & 71.98   \\ 
		RNTN(\cite{socher2013recursive}) & 55.56 & 45.45 & 50.00 && 77.78 & 84.00 & 80.77 && 72.22 & 80.76   \\ \hline\hline
		
		\rule{0pt}{3ex}\textbf{IMDB} &&&&&& \\
		\rowcolor{lightgray} Proposed method(CNN$+$Graph) & 87.42 & 90.85 & 88.31 && 86.25 & 86.80 & 86.60 && 86.07 & 87.27   \\
		SVM(linear) & 77.37 & 76.01 & 76.69 && 75.70 & 77.07 & 76.38 && 76.53 & 76.54   \\
		SVM(RBF) & 65.85 & 58.70 & 62.07 && 67.80 & 74.07 & 70.80 && 67.00 & 70.79   \\
		NB(tf-idf) & 74.72  &  73.41 &  74.06 &&  73.84 &  75.14 & 74.49  && 74.27  & 74.48   \\
		Kim(CNN$+$w2v) & 81.84 & 82.35 & 81.29 && 82.31 & 82.32 & 81.01 && 79.97 & 81.11   \\ 
		RNTN(\cite{socher2013recursive}) & 80.98 & 80.21 & 80.59 && 80.38 & 81.14 & 80.76 && 80.67 & 80.75   \\ 
			
		\bottomrule
	\end{tabular}
	} 
\end{table}

With the intention to show the priority of the graph representation procedure over word2vec, we have extracted the word embeddings only on the  IMDB dataset to demonstrate the effect of graph representation on text documents. The obtained results in Table \ref{tbl:cmp} designate that CNN trained on features extracted from limited corpus, performs weaker than the graph-based features and globally trained word embeddings. This shows the superiority of the graphs in extracting features from the text materials even if the corpus size is limited. It is worth mentioning that the word graphs are made only out of the available corpus and are not dependent on any external features. 

\begin{table}[ht]
	\caption{Comparison of graph-based learning vs. word2vec}
	\label{tbl:cmp}
	\begin{tabular}{lcccccccccc}
		\toprule
		
		\multirow{2}{*}{Method} &
		\multicolumn{3}{l}{Negative class (\%)}  &&
		\multicolumn{3}{l}{Positive class (\%)}  && 
		\multicolumn{2}{l}{Overall (\%)} \\ \cmidrule{2-4} \cmidrule{6-8} \cmidrule{10-11}
		
		& {precision} & {recall} & {F1} &&
		{precision} & {recall} & {F1}  &&
		{accuracy} & {F1} \\ 
		
		\midrule
		\rule{0pt}{3ex}\textbf{IMDB} &&&&&& \\
		Graph & 87.42 & 90.85 & 88.31 && 86.25 & 86.80 & 86.60 && 86.07 & 87.27   \\
		w2v & 74.34 & 73.37 & 75.20 && 71.41 & 70.82 & 71.32 && 70.14 & 72.71   \\ 
		
		\bottomrule
	\end{tabular}
\end{table}

\FloatBarrier
\subsection{Sensitivity Analysis}

The convolutional neural network offered for sentence classification can benefit from four models in terms of using the word vectors, namely, CNN-rand, CNN-static, CNN-non-static, and CNN-multichannel. In CNN-rand, all the word vectors are randomly initialized and then optimized through the training phase. CNN-static uses pre-trained word vectors, and for those without a vector (new or unknown words) the vector is randomly initialized. All the vectors are kept static and only the other parameters of the model are learned in the training phase. CNN-non-static is the same as CNN-static, but word vectors are optimized and fine-tuned. Finally,  the CNN-multichannel model uses two sets of word vectors each treated as a channel. In one channel, the word vectors (embeddings) are updated, in the latter they remain static.

We demonstrate the performance of the above described models on the sampled data from all available datasets (250 negative and 250 positive documents divided into train and test on 80-20 ratio) to figure out which model is the best choice to be coupled with graph embeddings. Table~\ref{tbl:cnn-model} presents the performance of different models on the sample data. The results reveal that the CNN-static model is close and at some levels is better than CNN-non-static. Moreover, this indicates that the feature set extracted from the graphs are rich enough and don't require optimization and fine-tuning.

\begin{table}[ht!]
	\caption{Comparison of different CNN models on the sampled data with graph embeddings}
	\label{tbl:cnn-model}
	\resizebox{1\textwidth}{!}{
	\begin{tabular}{lcccccccccc}
		\toprule
		
		\multirow{2}{*}{Method} &
		\multicolumn{3}{l}{Negative class (\%)}  &&
		\multicolumn{3}{l}{Positive class (\%)}  && 
		\multicolumn{2}{l}{Overall (\%)} \\ \cmidrule{2-4} \cmidrule{6-8} \cmidrule{10-11}
		
		& {precision} & {recall} & {F1} &&
		{precision} & {recall} & {F1}  &&
		{accuracy} & {F1} \\ 
		
		\midrule
		\rule{0pt}{3ex}\textbf{Sampled data} &&&&&& \\
		CNN-rand & 51.79 & 60.42 & 55.77 && 56.82 & 48.08 & 52.08 && 54.00 & 52.08 \\
		CNN-static & 64.29 & 52.94 & 58.06 && 58.62 & 69.39 & 63.55 && 61.00 & 63.55 \\ 
		CNN-non-static & 54.55 & 62.50 & 58.25 && 60.00 & 51.92 & 55.67 && 57.00 & 55.67 \\ 
		CNN-multichannel & 52.17 & 51.06 & 51.61 && 57.41 & 58.49 & 57.94 && 55.00 & 57.94 \\ 
		
		\bottomrule
	\end{tabular}
	} 
\end{table}

\begin{figure}[!ht]
	\centering
	\includegraphics[width=0.85\textwidth]{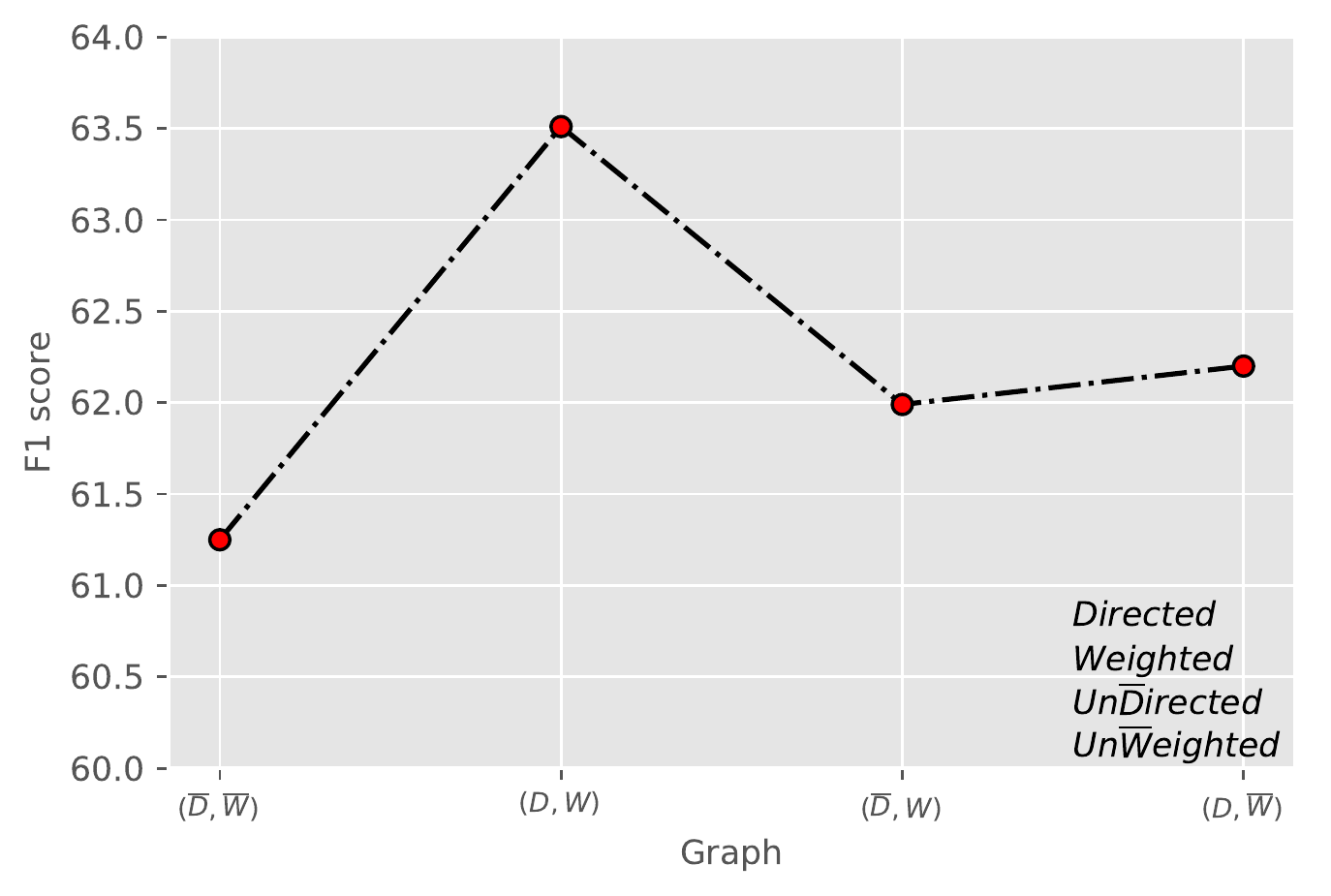}
	\caption{Sensitivity of the proposed model to different graphs (directed, undirected, weighted, unweighted) in the proposed method.}
	\label{fig:graph}
\end{figure}

Four types of graph can be used in the proposed method, directed weighted, undirected weighted, directed unweighted, and undirected unweighted. Each of these graphs has its own specific features and can show different characteristics of a text. As an example, the directed graph can represent the order of words in a sentence, while weights in a graph can represent how often words appear together in the text. 
Figure \ref{fig:graph} displays the model sensitivity to various graphs. As it is shown in Figure \ref{fig:graph}, the directed weighted graph results in better performance and that's why it is used for the experiment.

\begin{figure}[!ht]
	\centering
	\includegraphics[width=0.85\textwidth]{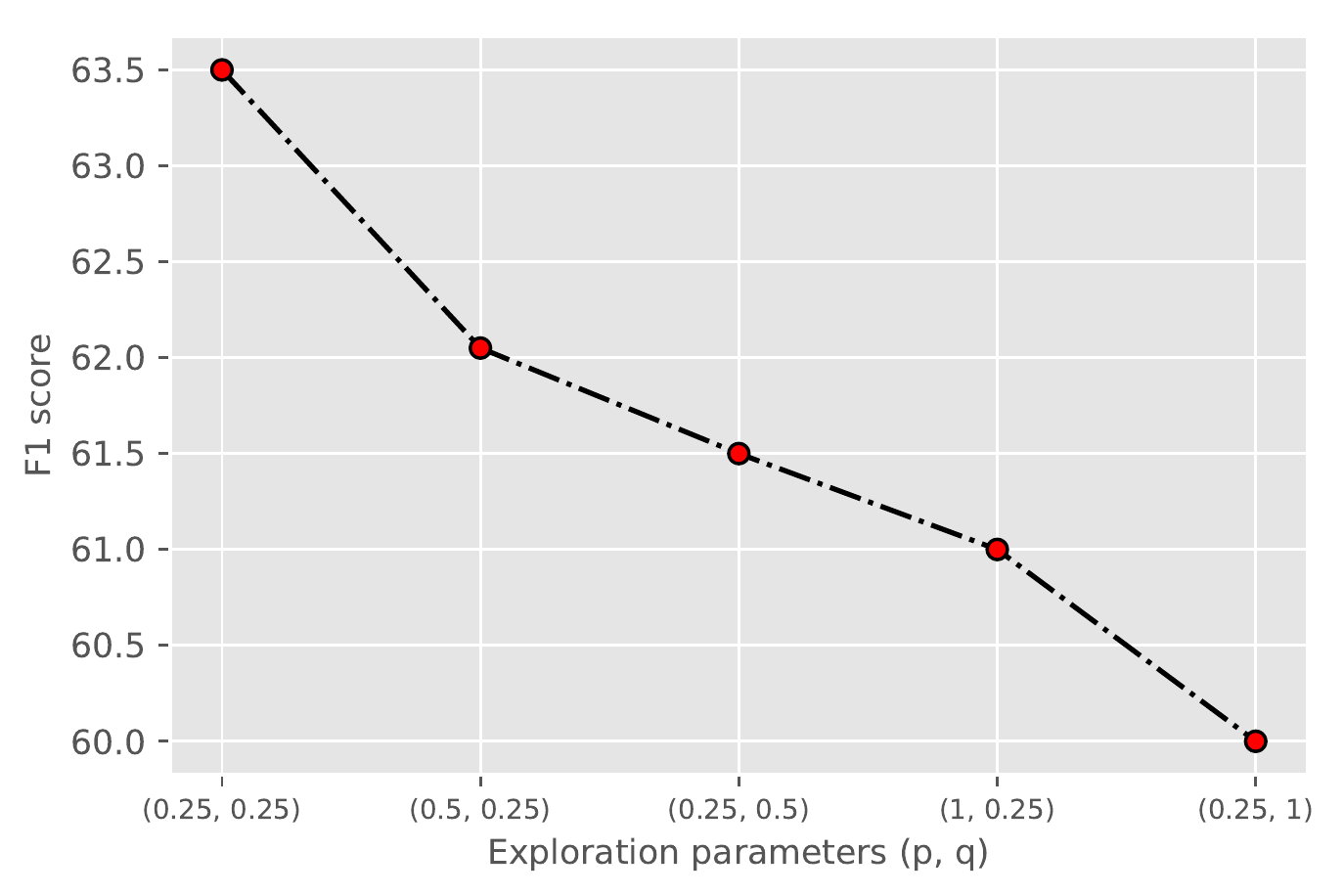}
	\caption{Sensitivity analysis for the $p$ and $q$ values of the proposed algorithm over sampled data from all of the datasets.}
	\label{fig:sens}
\end{figure}

The proposed method relies on two main parameters for graph exploration, $p$ and $q$. We examine how the different choices of parameters affect its performance (F1) the sampled data. As can be seen from Figure~\ref{fig:sens}, the performance is high for low values of $p$ and $q$. While a low q encourages outward exploration of the graph, it is balanced by a low value of $p$ which ensures the random walk not to go enormously far from the starting node.

\begin{figure}[!ht]
	\centering
	\includegraphics[width=0.85\textwidth]{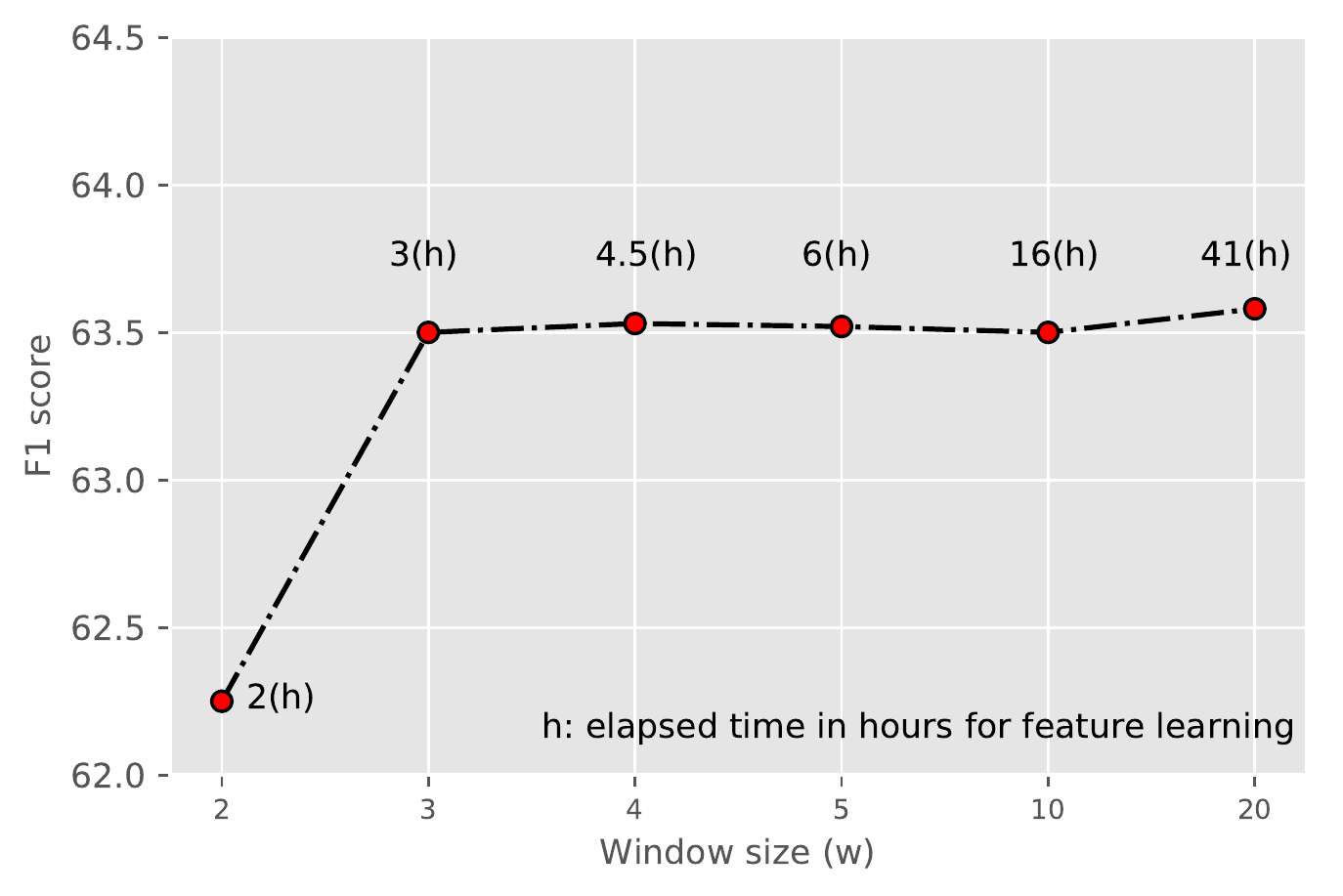}
	\caption{Performance analysis of the proposed algorithm for different window sizes over the sampled dataset.}
	\label{fig:window}
\end{figure}

In the process of graph formation from the given documents, the window size highly impacts the performance of the algorithm. Based on a variety of experiments, a window size of 2 to 10 seems to be relevant to make the word graphs. However, high values result in a very dense graph which takes a lot of processing time to be transformed into features and low values result in a low-quality feature set. Our analysis, Figure~\ref{fig:window}, shows that a word window of size 3 is an appropriate choice in terms of runtime and accuracy. 

\FloatBarrier
\section{Conclusion and Future Work}
\label{sec:conclusion}
\textcolor{black}{In this paper, we proposed a new graph representation learning approach for textual data. By incorporating different hidden aspects in a graph-based text representation, the proposed framework succeeded to incorporate most of the features in documents for the polarity identification task. The unsupervised graph representation learning was applied to extract the continuous and latent features to employ them in learning schema. The experimental results affirmed the superiority of the proposed method versus its competitors. Furthermore, deep learning architectures were employed to demonstrate the strength of the proposed method on the sentiment classification. The graph structure enabled the proposed framework to incorporate the stop words, word positions, and more importantly word orders as opposed to the traditional techniques. The obtained results on standard datasets have verified the usefulness of graph-based representation aligned with deep learning based on the performance improvements in the sentiment classification task. This ongoing field of research has several directions that could be followed for the future practice, including, but not limited to, employing other graph-based representation methods to extract hidden characteristics of a network, exploiting preprocessing methods to enrich the initial features of the network, and employing other innate features and informations in the social media to enhance sentiment analysis techniques.}

\section{Acknowledgments}
The authors would like to thank the anonymous reviewers for contributing valuable comments and recommendations which significantly intensified the quality of the paper.

\section*{References}

\bibliographystyle{model5-names}
\biboptions{authoryear}

\bibliography{ref.bib}

\end{document}